\documentclass{article}

    \PassOptionsToPackage{numbers, compress}{natbib}

    \usepackage[preprint]{neurips_2025}

\usepackage[utf8]{inputenc} 
\usepackage[T1]{fontenc}    
\usepackage{hyperref}       
\usepackage{url}            
\usepackage{booktabs}       
\usepackage{amsfonts}       
\usepackage{nicefrac}       
\usepackage{microtype}      
\usepackage{xcolor}         
\usepackage{amsmath}
\usepackage{graphicx}
\usepackage{subcaption}

\title{RealCam-Vid: High-resolution Video Dataset with Dynamic Scenes and Metric-scale Camera Movements}

\author{%
  Guangcong Zheng$^*$, Teng Li$^*$, Xianpan Zhou\thanks{Equal Contribution}, Xi Li
  \\
  Department of Computer Science\\
  Zhejiang University\\
  \texttt{guangcongzheng@zju.edu.cn} \\
}

\begin{document}

\maketitle

\begin{abstract}
Recent advances in camera-controllable video generation have been constrained by the reliance on static-scene datasets with relative-scale camera annotations, such as RealEstate10K . While these datasets enable basic viewpoint control, they fail to capture dynamic scene interactions and lack metric-scale geometric consistency—critical for synthesizing realistic object motions and precise camera trajectories in complex environments . To bridge this gap, we introduce the first fully open-source, high-resolution dynamic-scene dataset with metric-scale camera annotations in \url{https://github.com/ZGCTroy/RealCam-Vid}.
\end{abstract}

\begin{figure}[htbp]
    \centering
    \includegraphics[width=\linewidth]{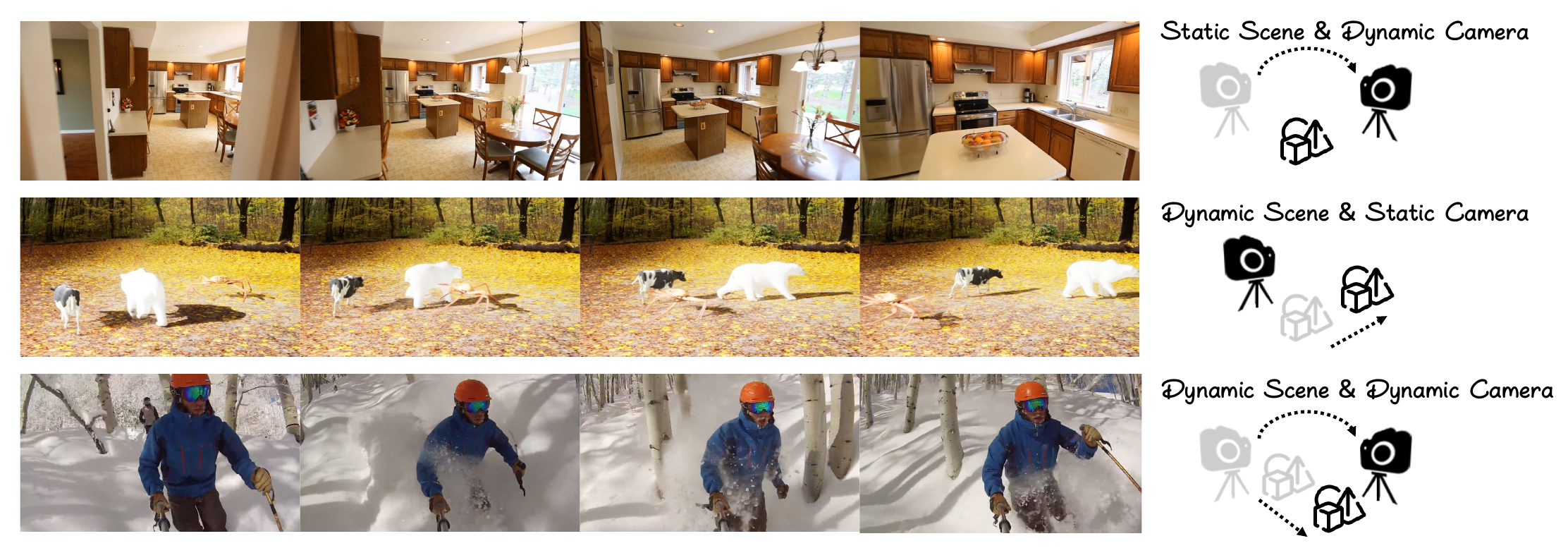}
    \caption{
        \textbf{Overview of Existing Datasets for Camera Motions and Scene Dynamics.}
        Static Scene \& Dynamic Camera videos boasts high aesthetic quality with dense relative-scale camera trajectory annotations but lacks object dynamics, which may lead to overfitting on rigid structures. Dynamic Scene \& Static Camera videos capture dynamic objects yet omit camera motion, limiting their applicability in trajectory-based video generation. Dynamic Scene \& Dynamic Camera videos feature rich real-world dynamics with both moving objects and camera motion while lack metric-scale camera annotations, rendering them unsuitable for metric-scale training. In this technical report, we release the first open-sourced high-resolution video dataset with dynamic scenes and metric-scale camera parameters in \url{https://github.com/ZGCTroy/RealCam-Vid}.
    }
    \label{fig:caption}
\end{figure}

\section{Technical Report}

\begin{figure}[htbp]
    \centering
    \includegraphics[width=\linewidth]{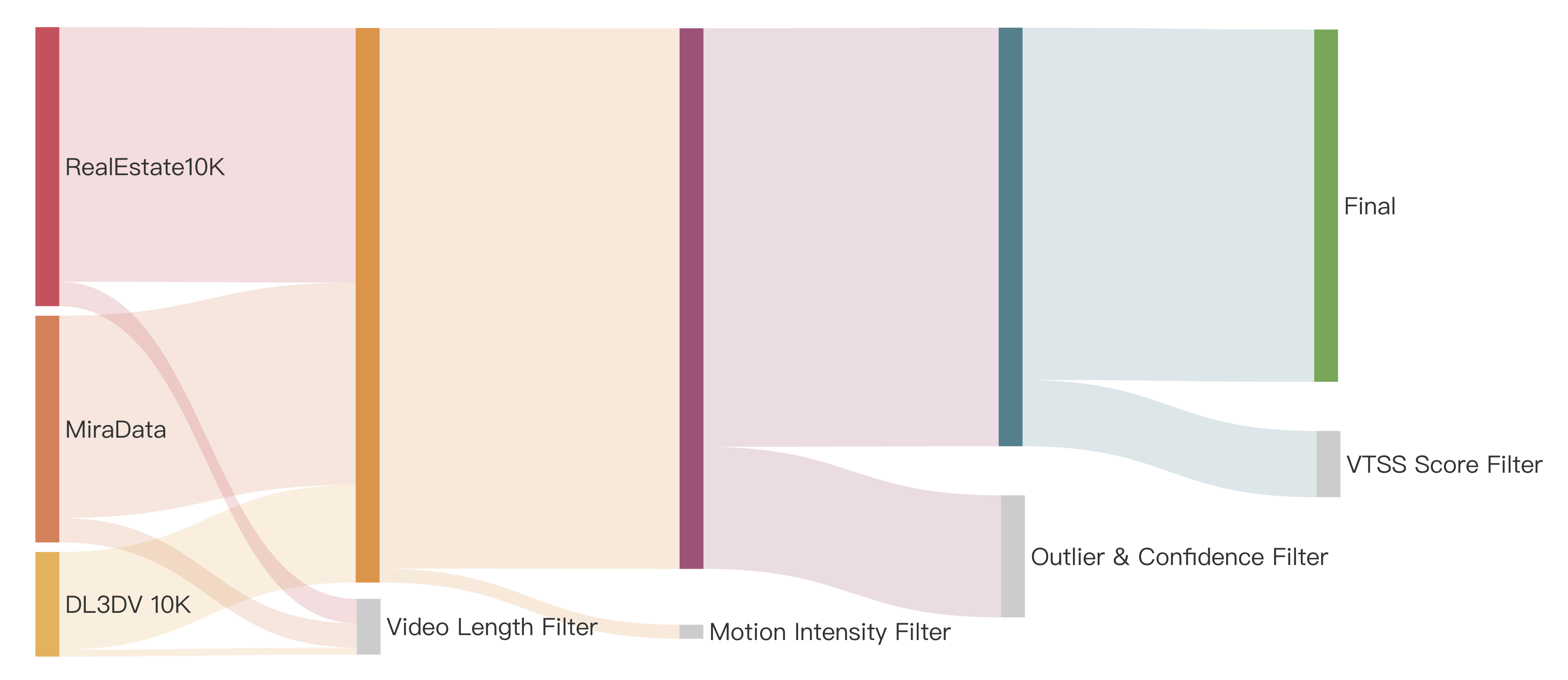}
    \caption{
        \textbf{Our Data Filtering Pipeline.}
        We employ a series of filters to refine the dataset, starting with three distinct sources: RealEstate10K~\cite{zhou2018stereo}, MiraData~\cite{ju2024miradata}, and DL3DV-10K~\cite{ling2024dl3dv}. These datasets undergo a series of stages, with key filters applied, including Video Length, Motion Intensity, and Outlier \& Confidence filters. The final dataset, after processing through these filters, is curated using the VTSS Score Filter from Koala-36M~\cite{wang2024koala}. Gray bars show the amount of data filtered out by each filter, while the colored bars indicate the remaining data at each stage.
    }
    \label{fig:caption}
\end{figure}

Current datasets for camera-controllable video generation face critical limitations that hinder the development of robust and versatile models. Our curated dataset and data-processing pipeline uniquely combines diverse scene dynamics with metric-scale camera trajectories, enabling generative models to learn both scene dynamics and camera motion in a unified framework.

\subsection{Video Clip Splitting}

Video clip splitting plays a pivotal role in ensuring temporal coherence and physical plausibility of camera motion trajectories in generated videos.
Traditional scene-cut detection methods often fail to distinguish gradual transitions (e.g., fades, slow tracking shots), leading to discontinuous motions. For instance, misaligned splits disrupt optical flow consistency, causing unnatural ``jumps'' in camera trajectories.
To partition video sequences into semantically coherent segments, we adopt the split operator proposed in Koala-36M~\cite{wang2024koala}, a data-driven approach leveraging pretrained spatiotemporal representations to identify scene boundaries by analyzing temporal coherence in feature embeddings extracted from video clips. Specifically, it computes similarity scores between consecutive frames and detects split points where scores fall below an adaptive threshold derived from global statistics.

This method significantly outperforms traditional libraries like PySceneDetect, which relies on handcrafted thresholds for histogram or edge-based dissimilarity metrics. While PySceneDetect often fails in dynamic scenes with gradual transitions or lighting variations, Koala-36M’s learned embeddings inherently capture contextual and motion cues, enabling robust detection of both abrupt and smooth transitions. 
To ensure temporal continuity and sufficient camera motion consistency for downstream tasks, we implement a clip-length filtering criterion: Video clips containing fewer than 49 frames are systematically excluded from the dataset.

\subsection{Motion Intensity Filtering}

Video sequences containing static camera movements (i.e., minimal viewpoint changes) pose challenges for camera-controlled video generation training. These static video clips typically provide insufficient motion priors for neural networks to learn meaningful camera motion patterns, potentially leading to model degeneration where the generator defaults to static camera shots regardless of motion instructions.
Static sequences often contain inherent noise that becomes perceptually amplified when processed by motion-sensitive generative models, resulting in visual artifacts.
To identify and filter such suboptimal sequences, we adopt a keypoint trajectory analysis approach inspired by VBench-I2V~\cite{huang2024vbench}. Specifically, we employ CoTracker~\cite{karaev2024cotracker}, a state-of-the-art video correspondence estimator, to track keypoints across consecutive frames. 
The motion magnitude is quantified by calculating the average displacement of keypoints.
Empirical analysis on our training corpus revealed that sequences with motion threshold \(5\% \times \min(H,W) \) predominantly exhibit static camera characteristics, effectively capturing subtle background shifts while excluding clinically static content.

\subsection{Long Caption and Short Caption}


Building on insights from previous works (e.g., PixArt~\cite{chen2023pixart} and DALL-E 3~\cite{betker2023improving}), which emphasize the critical role of caption granularity in generative models, we adopt CogVLM2-Caption~\cite{yang2024cogvideox} to address the limitations of existing video-text datasets. Unlike conventional approaches that generate oversimplified or noisy captions, CogVLM2-Caption leverages a hybrid architecture to produce structured long-form descriptions with spatiotemporal coherence, enabling precise alignment between video content and textual metadata.
The model preserves inter-frame dependencies and allows dynamic weighting of key visual elements across time steps, effectively capturing gradual scene transitions through temporal context aggregation.

\subsection{Camera Annotation for Dynamic Scenes}

Our pipeline prioritizes robust motion-aware processing method, MonST3R~\cite{zhang2024monst3r}, for reliable pose estimation on videos with dynamic scenes.
Unlike COLMAP~\cite{schoenberger2016sfm}, which rely on keypoint matches vulnerable to dynamic outliers, this state-of-the-art method explicitly models per-frame geometry while distinguishing moving objects from static scenes.
This method automatically initializes region masks through either RAFT~\cite{teed2020raft} derived motion cues or SAM~\cite{ravi2024sam} generated semantic segmentation, enabling simultaneous refinement of scene geometry separation and ego-motion computation. 
Camera parameters are subsequently recovered through constrained optimization that enforces geometric consistency across static regions, with dynamic elements marginalized via uncertainty-aware cost functions. 
This integrated approach achieves temporal pose coherence without requiring explicit per-object motion modeling, significantly reducing annotation dependencies while handling complex real-world motion patterns.

For global optimization, we optimize the MonST3R loss for 300 iterations with a linear scheduler.
We skip the first 10\% of optimization steps and start supervising the flow loss only when the extracted camera poses achieve rough alignment, with the average value below 25.
We empirically set \(w_{\rm flow}=0.01\) and \(w_{\rm smooth}=0.01\) for better prediction accuracy.
To balance accuracy and computational efficiency, we employed a strided sliding window strategy on the scene graph, with windows size 9 and stride 2.

\begin{figure}[htbp]
    \centering
    \includegraphics[width=\linewidth]{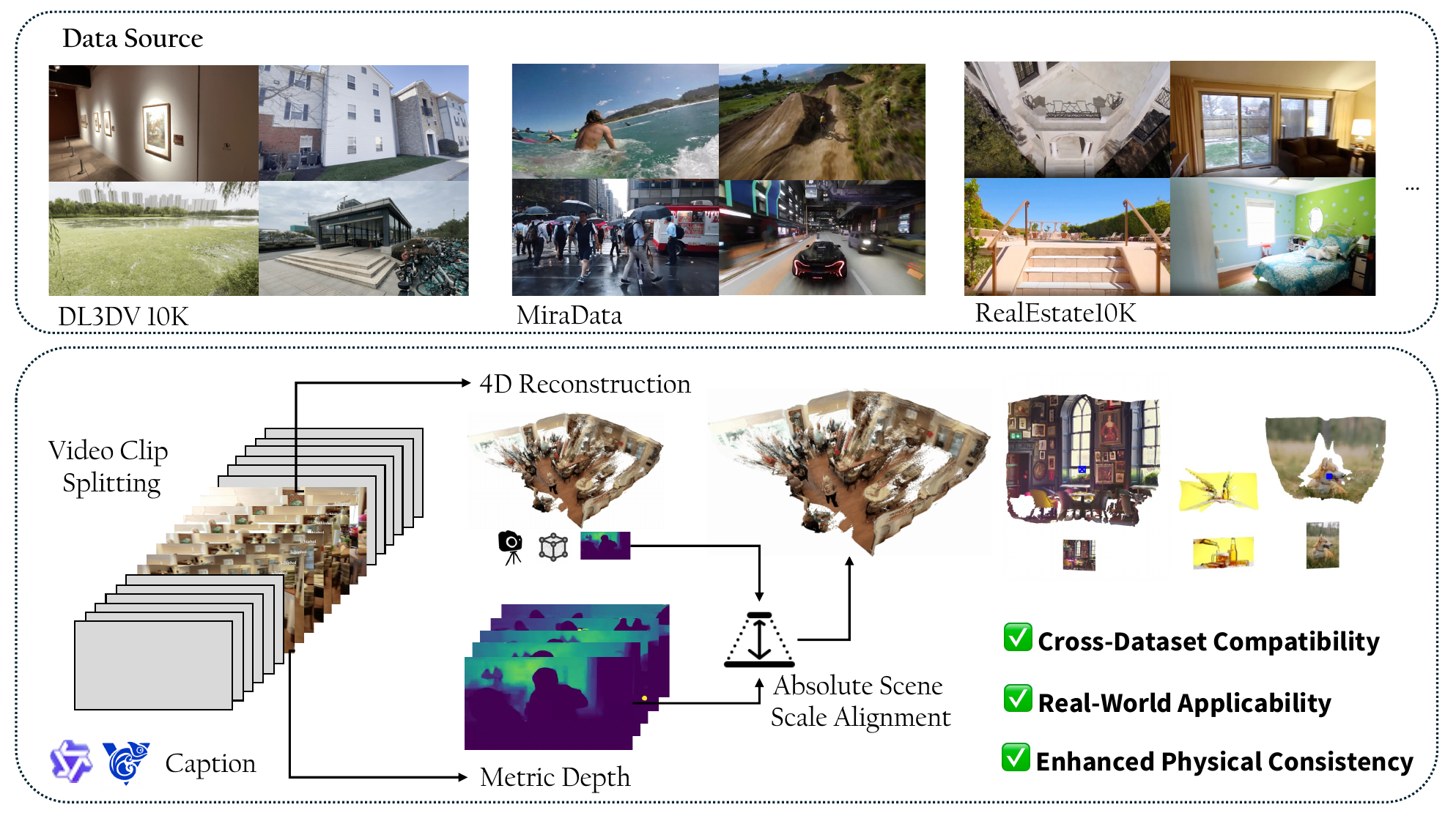}
    \caption{
        \textbf{Pipeline for Metric Scale Alignment.}
        This diagram illustrates the process of calibrating heterogeneous video sources to achieve cross-dataset compatibility by aligning relative-scale camera trajectory to absolute, metric scales. Depth maps are converted to disparity maps to suppress distant noise and highlight near-field detail. Metric-scale estimates are obtained via a metric depth predictor, while relative-scale disparities come from 4D reconstructions. 
    }
    \label{fig:caption}
\end{figure}

\subsection{Metric Alignment for A Unified Scene Scale}

Accurate scale alignment of camera trajectories emerges as a fundamental requirement when constructing 3D vision datasets from heterogeneous video sources. These datasets inherently exhibit divergent scale definitions, where a "unit length" in one source (e.g., normalized coordinates) may not correspond to physical measurements in another (e.g., sensor-calibrated sequences). Without explicit metric grounding, learned models inherit dataset-specific scale biases, compromising their capacity to generalize to real-world physical constraints.

We calibrate camera trajectories by utilizing depth maps to align relative scales to absolute metric scales. 
We first convert the depth maps to disparity maps to suppress distant noise and enhance near-field details.
For a \(N\)-frame video sequence \(\{{\bf V}_i\}_{i=1}^N\), the metric-scale disparity estimates \(\{{\bf D}^{abs}_i\}_{i=1}^N\) are obtained for each frame using a metric depth predictor (e.g., Metric 3D~\cite{hu2024metric3d}), while the relative-scale disparity values \(\{{\bf D}^{rel}_i\}_{i=1}^N\) are derived from structure-from-motion results (by e.g., COLMAP~\cite{schoenberger2016sfm}). The scale factor for scene-level metric scale alignment can thus be formulated as:

\begin{equation}
    s = \mathop{\arg\min}_s \sum_{1\le i \le N} \left\| {\bf D}^{abs}_i - s \cdot {\bf D}^{rel}_i \right\|_2^2
\end{equation}

To enhance numerical stability during scale factor computation, we implement a disparity value masking strategy that discards measurements in the near/far planes (defined as the top and bottom 5\% disparity ranges, corresponding to potential noise regions) while exclusively selecting disparity values with confidence scores within the top 50\% percentile.
We solve for the optimal scale factor \(s^*\) that minimizes the L2-norm residual by the least square method:

\begin{equation}
    s^* = \frac{\sum_{i=1}^N{{\bf D}^{abs}_i \cdot {\bf D}^{rel}_i}}{\sum_{i=1}^N{({\bf D}^{rel}_i)^2}}
\end{equation}

The final scale factor \(s^*\) is applied to the translation vector \({\bf t} \in {\mathbb R}^{3\times1}\) from the per-frame extrinsic matrix of relative-scale camera trajectories before alignment, obtaining the per-frame metric-scale extrinsic matrix \({\bf E} = \begin{bmatrix} {\bf R}, s^* \cdot {\bf t} \end{bmatrix} \in {\mathbb R}^{3 \times 4}\), where \({\bf R} \in {\mathbb R}^{3 \times 3}\) is the rotation matrix.

{
    \small
    
    \bibliographystyle{abbrv}
    \bibliography{main.bib}
}

\end{document}